# Generative AI for learning: Investigating the potential of synthetic learning videos

Daniel Leiker[1,3][0000-0002-8438-9185], Ashley Ricker Gyllen[2,3][0000-0002-5549-2247], Ismail Eldesouky[3][0000-0001-5084-3557], and Mutlu Cukurova[1][0000-0001-5843-4854]

[1] UCL Knowledge Lab, London WC1N 3QS, UK
[2] Metropolitan State University, Denver CO 80204, USA
[3] EIT InnoEnergy, Eindhoven, NL
`daniel.leiker.16@ucl.ac.ui`

**Abstract.** Recent advances in generative artificial intelligence (AI) have captured worldwide attention. Tools such as Dalle-2 and ChatGPT suggest that tasks previously thought to be beyond the capabilities of AI may now augment the productivity of creative media in various new ways, including through the generation of synthetic video. This research paper explores the utility of using AI-generated synthetic video to create viable educational content for online educational settings. To date, there is limited research investigating the real-world educational value of AI-generated synthetic media. To address this gap, we examined the impact of using AI-generated synthetic video in an online learning platform on both learners' content acquisition and learning experience. We took a mixed-method approach, randomly assigning adult learners ($n$ = 83) into one of two micro-learning conditions, collecting pre- and post-learning assessments, and surveying participants on their learning experience. The control condition included a traditionally produced instructor video, while the experimental condition included a synthetic video with a realistic AI-generated character. The results show that learners in both conditions demonstrated significant improvement from pre- to post-learning ($p < .001$), with no significant differences in gains between the two conditions ($p = .80$). In addition, no differences were observed in how learners perceived the traditional and synthetic videos. These findings suggest that AI-generated synthetic learning videos have the potential to be a viable substitute for videos produced via traditional methods in online educational settings, making high quality educational content more accessible across the globe.

## 1 Introduction

The argument for using artificial intelligence (AI) to support learning and education is well established, with a growing body of evidence demonstrating the positive impacts of using AI to support learning, engagement, and metacognitive development [1, 2, 3, 4]. However, generative AI is a relatively new area with respect to its implementation in learning contexts, and the extent to which AI-generated media can be used to support human learning remains largely unexamined. Recent advances in generative AI have captured worldwide attention. Tools such as Dalle-2 and ChatGPT, developed by OpenAI, suggest that tasks previously thought to be beyond the capabilities of AI may now augment the productivity of creative media and educational content in various new ways. Since at least 2014, new methods in generative machine learning, such as Generative Adversarial Networks (GANs), have enabled the realistic synthesis of digital content [5]. Over time, these models are increasing in size and complexity resulting in greater sophistication of their generated outputs [6], including the generation of photo-realistic images, cloning of voices, and animation of faces [7, 8, 9, 10]. More recently, methods such as Generative Pre-trained Transformers (GPT) are ushering in a new era of generative AI capability [11, 12]. Generative AI technologies like these are already being leveraged across several industries including entertainment, customer services and marketing [13]. Their introduction at scale can help us address global educational challenges such as access to high quality content across the globe and make progress towards global sustainable development goals (e.g., SDG4). In addition, they have the potential to reshape our creative and knowledge-based workforces, improve online learning, and transform large sectors of our economies.

The global demand for massive open online courses (MOOCs), online degree and training programs, and employee upskilling and reskilling is growing rapidly. This is particularly true for low- and middle-income countries, where access to well-designed, high quality educational resources is a major problem. This demand in turn drives the need for educational content for these online platforms, including a significant amount of instructional learning videos requiring periodic updates to keep up with trends and rapid innovation in research and technology. The purpose of instructional videos in online learning content is to enhance the pedagogy, or message [14]. According the to the cognitive theory of multimedia learning, for these instructional learning videos to be effective, they should be designed with human cognition in mind [15]. Additionally, they should not display unnecessary, excessive, extraneous elements (e.g., overabundance of motion graphics) that can distract from learning and overwhelm cognitive load [16]. Taking an evidence-based approach to creating synthetic learning videos using generative AI is an appealing way to meet these needs due to the challenges associated with producing high quality video media (e.g., lack of on-screen experience of the instructors, significant amount of time and resources needed, and relatively under-resourced nature of educational institutions and schools). For this study, we will specifically focus on the use of generative AI tools to create synthetic videos with virtual instructors, resembling traditional lecture videos found in online learning experiences.

## 2      Background Research and Context

Virtual instructors (or animated pedagogical agents) are lifelike onscreen characters, enacted by a computer to support learning by providing guidance or instruction through an online learning experience [17, 18]. Previous research demonstrates that including an animated pedological agent can improve learning in online settings [19, 20, 21, 22]. Similarly, multiple studies have shown that the addition of a character to virtual learning can positively impact learners' behaviors, attitudes, and motivation [23]. Given the recent advancements in generative AI, a logical next step in this line of research is to examine whether AI-generated virtual instructors can effectively support online learning [24]. To date, we know of one study investigating the use of an AI-generated virtual instructor to support human learning. In that study, researchers compared two different AI-generated characters and found that character likeability influenced participants motivation towards learning [25]. While this is a promising finding, research comparing learning from an AI-generated virtual instructor to learning from a traditional instructor is needed to evaluate the real-world educational value of such AI-generated media. To the best of our knowledge, this is the first study to make that comparison.

The experiment performed for this study was done in the context of online professional learning and in collaboration with EIT InnoEnergy, a European company promoting innovation and entrepreneurship in the fields of sustainable energy. InnoEnergy is part of the European Institute of Innovation and Technology which is itself a body of the European Union. They are spearheading efforts to decarbonize Europe through the leadership of the European Battery Alliance (EBA) Academy. The subject matter of the content used for this study was sampled from an introductory lesson from InnoEnergy's EBA Academy, which was designed using the basic principles of multimedia learning (e.g., contiguity, modality, coherence, segmenting, pre-training, practice) to support an effective learning experience [16].

The key audience for the introductory lesson we sampled from is technicians seeking employment in gigafactories producing lithium-ion batteries, engineers (e.g., electrical, chemical) looking to increase their knowledge in battery technology, and knowledge workers (e.g., upper-level managers, investors) looking to expand their knowledge of the battery industry for strategic decision making. The aim of courses like this one is to achieve the goal of accelerating workforce transitions toward a clean energy economy. One significant challenge in generating learning products and services to achieve this aim is that much of the content is in new and emerging fields (such as state-of-the-art battery manufacturing), where little to no prior content exists to draw from. Another challenge is the rapid pace at which research and technology in these industries are transforming, requiring fast and frequent iterations to certain subdomains of the curriculum sometimes as frequently as a couple of times a year. One potential solution to address these challenges is to apply the research that exists in the literature around multimedia learning [e.g., 15, 17, 19, 21] to the creation of new asynchronous online learning content while exploring the practical use of AI-generated synthetic videos as an alternative to traditional production methods. This study will help to determine the viability of this approach with the end goal of ensuring that the learner experience will be enhanced through these efforts. More specifically, in this research paper, we propose to address the following research questions.

   1) To what extent does the use of AI-generated synthetic videos in an online learning platform differentially impact learning performance when compared to traditionally produced instructor videos?

   2) What are the perceived differences between AI-generated synthetic videos and traditionally produced instructor videos for learners in an online educational setting?

## 2      Methodology

**Participants.** Our sample included 83 adult learners, recruited from a global professional learning community, ranging in age from 18 to 64 with an average age of 41.5 years. Of these learners, 73% identified as male, 19% identified as female, 0% identified as non-binary, and 8% preferred not to disclose their gender identity. With regards to education, 4% held an associate degree, 15% held a bachelor's degree, 58% held a master's degree, and 23% held a doctorate degree. Additionally, 68% identified as being unfamiliar with the subject matter prior to completing the micro-learning course, while 32% reported having prior knowledge of the subject matter.

**Synthetic Video Creation.** The key focus of this experiment is the introduction of AI-generated synthetic video used as instructional video content in a micro-learning course. An instructor video, produced using traditional recording methods, was used as both a control for our experiment and as the source material for generating the synthetic videos. To create the video for our experimental condition, the AI video creation platform Synthesia was used to generate text-to-videos (TTV) with photo-realistic synthetic actors. To establish the photo realistic quality for these generated videos Synthesia first records live footage of a particular training data of a live actor and then establishes a synthetic clone of that actor. Neural video synthesis is then applied to add realistic gestures and movements to the synthetically generated video representation of the actor initiated through TTV input for the final production asset. This process could be applied to any "actor" or "instructor" to create a synthetic video clone of themselves for repurposing in this type of AI-generated synthetic video creation. The AI-generated character used in this study was matched to the instructor as closely as possible with respect to age, gender, and race. This resulted in two videos with identical instructional content, but with different visual representations of the instructor (see Figure 1).

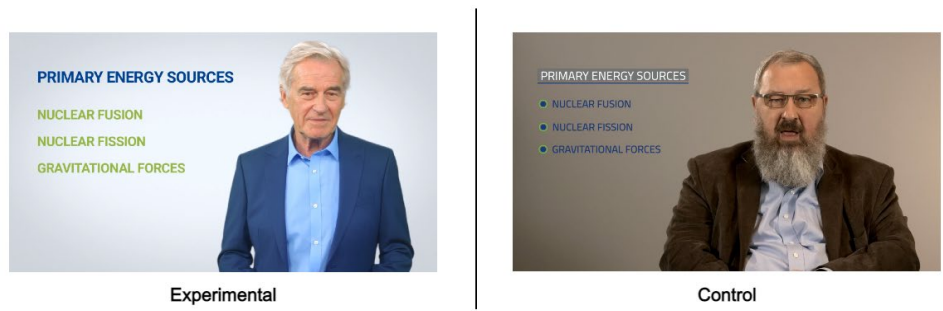

**Fig. 1.** Screenshots of the AI-generated virtual instructor in the experimental condition (left) and the traditional instructor in the control condition (right)

**Learning Design.** Using the two videos described above, two micro-learning courses were designed for this experiment on the topic of energy sources and vectors at an introductory level. The main goal of the micro-course was to provide learners with an easy-to-complete learning unit that could be accessed on a mobile device and centered around the instructional video content. The micro-course consisted of a series of activities, including a course introduction page, a pre-learning knowledge check, the instructional video content (4.5 minutes in length), a click and reveal application activity, and a post-learning assessment. The courses were delivered to learners via EdApp a learning platform designed with a focus on both mobile users and supporting micro-learning formats. They were identical except for the instructional video content (i.e., the control condition included the traditional instructor video, while the experimental condition featured the synthetic video).

**Procedure.** We utilized a mixed-method approach to this research, by collecting and examining both quantitative and qualitative data. Participants were invited through an email campaign by EIT InnoEnergy via their mailing list for current students and alumni of their master's degree and EBA Academy programs, as well as their employees engaging in learning and development. All members of this mailing list had previously consented to receive promotional emails for research purposes. Upon clicking the received promotional email link, participants were directed to sign into the EdApp platform. After completing informed consent, nimble links was used to randomly assign participants into either the experimental or control condition, they completed the micro-learning course described above, and were then asked to complete a survey on their learning experience. As part of the survey, participants used a 5-point Likert Scale to indicate their level of agreement with the following questions:

*1) I would consider my overall experience with this micro-learning course positive.*
*2) The use of video in the course met my expectations.*
*3) The use of video in the course improved my understanding of the material.*
*4) I would be interested in taking other courses like this.*

Additionally, participants were asked to respond to the open-ended question: *Do you have any overall suggestions for improving this course?*

## 3   Results

To evaluate Research Question 1, responses to the pre- and post-learning assessments were scored and a difference score (post minus pre) was calculated to represent knowledge gains. To confirm the appropriateness of proceeding with a parametric test, skew and kurtosis of the difference scores were estimated and the data were visually inspected for normality. Subsequently, a paired sample t-test was used to compare pre- and post-learning across the entire sample. Regression analyses were used to test if the condition (experimental vs. control) significantly predicted knowledge gains. Regression analyses were then used to examine differences between conditions, as they allowed us to control for prior subject matter knowledge. To evaluate Research Question 2, quantitative and qualitative approaches were used to examine participant responses to the learning experience survey. For close-ended questions with Likert response options, Pearson Chi-Square analyses were used to examine differences between conditions. For the open-ended question, automated sentiment and thematic analyses were used to summarize the results and compare the findings between conditions. Automated coding was then cross-checked manually by researchers who co-authored the paper. The analyses for all quantitative data were completed using relevant packages in R, and the analyses for all qualitative data were completed using NVivo.

### 3.1 Impacts of AI-generated synthetic videos on learning performance

Based on paired sample t-tests, learners showed significant improvement from pre-learning ($M = 0.53$, $SD = 0.65$) to post-learning ($M = 1.51$, $SD = 1.03$) across the full sample of 83 participants, $t(82) = 8.31$, $p < .001$, $d = 0.91$. Demonstrating that the micro-learning course was effective at facilitating gains in content knowledge. Regression analyses indicated that condition (experimental vs. control) was not a significant predictor of knowledge gains ($\beta = .03$, $p = .80$, $r = .03$). This finding was unchanged when controlling for participants' pre-learning performance ($\beta = -.03$, $p = .79$, $r = .03$) or their self-reported prior knowledge ($\beta = .01$, $p = .92$, $r = .01$). This suggests that there was no significant difference in knowledge gains for participants who viewed the AI-generated synthetic video ($M_{gains} = 1.00$, $SD = 1.04$, $n = 52$) compared to participants who viewed the traditional instructor video ($M_{gains} = 0.94$, $SD = 1.13$, $n = 31$). The change in learning performance from pre- to post-learning in both conditions as well as across the full sample are presented in Table 1.

**Table 1.** Learner Performance and Knowledge Gains from Pre to Post

|  | Pre-Learning M (*SD*) | Post-Learning M (*SD*) | Knowledge Gains M (*SD*) | p (*d*) |
|---|---|---|---|---|
| *Experimental (n = 52)* | 0.45 (*0.61*) | 1.45 (*0.95*) | 1.00 (*1.04*) | < .001 *(0.96)* |
| *Control (n = 31)* | 0.66 (*0.70*) | 1.59 (*1.16*) | 0.94 (*1.13*) | < .001 *(0.83)* |
| ***Full Sample (n = 83)*** | **0.53 (*0.66*)** | **1.51 (*1.03*)** | **1.03 (*1.11*)** | **< .001 *(0.91)*** |

### 3.2 Learner perceptions of AI-generated synthetic videos

**Quantitative Findings.** Of the full sample ($n = 83$), 80% provided agreement ratings for the close-ended statements. Pearson Chi-Square analyses indicated that there were no significant differences in responses to the four close-ended questions between the experimental and control conditions (all $p > .46$). Specifically, agreement frequency was not significantly different between conditions for the following statements: I would consider my overall experience with this micro-learning course positive ($\chi^2(3) = 0.19$, $p = .98$, $V = .05$); The use of video in the course met my expectations ($\chi^2(4) = 4.65$, $p = .34$, $V = .27$); The use of video in the course improved my understanding of the material ($\chi^2(4) = 2.59$, $p = .63$, $V = .20$); I would be interested in taking other courses like this ($\chi^2(3) = 0.08$, $p = .99$, $V = .03$). Figure 2 presents the percentage of participants who agreed or strongly agreed with each statement for both conditions and the full sample. These findings suggest that participants who viewed the AI-generated synthetic video were just as likely as participants who viewed the traditional instructor video to agree that their overall learning experience was positive, that the video met their expectations, that the video improved their understanding of the content, and that they would be interested in taking other micro-learning courses like the one they took.

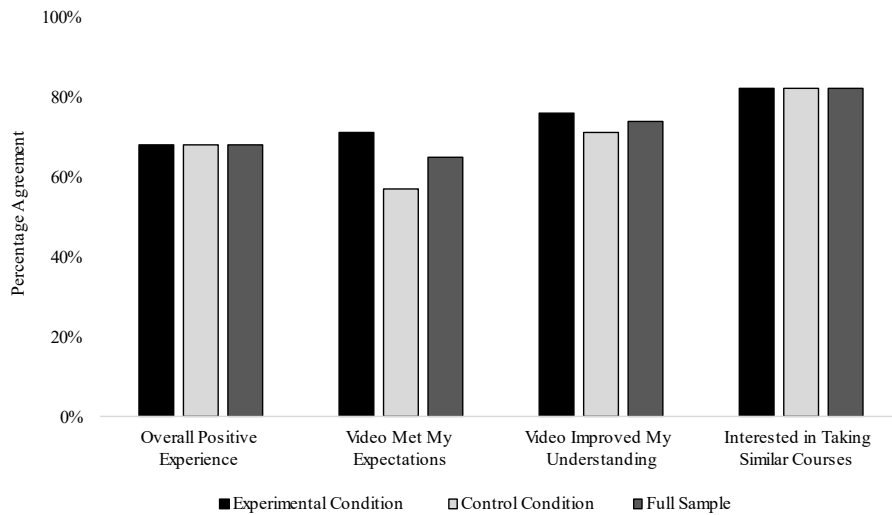

**Fig. 2.** Percentage of participants that agreed with each of the close-ended statements broken down by condition and across the full sample.

In addition to the lack of significant difference in agreement frequency between the experimental and control conditions, the pattern of Likert responses was identical for three of the four statements as well. Only the statement *The video in the course met my expectations* displayed a different pattern of responses between the experimental and control condition. That is, in the experimental condition participants were more likely to strongly agree with this statement than to remain neutral, while in the control conditions, participants were more likely to remain neutral than to strongly agree. Although this difference was not significant with the current sample size, the effect size indicator (*Cramer's V* = .27) suggests a medium effect.

**Qualitative Findings.** For the open-ended survey question asking for overall suggestions for improving this course, 20% of the sample provided a response. This response rate was similar for participants in the experimental condition (21%) and control condition (18%). When completing the survey, participants were unaware of which condition, they were randomly assigned into, and some of the participants in the AI condition did not realize the video was synthetic. For example, one participant in the AI-generated condition responded to the open-ended question with *"I do not see where the AI Content generated comes in?"*. A qualitative sentiment analysis revealed that 47% of responses displayed negative sentiment, 37% displayed neutral sentiment, and 16% displayed positive sentiment. Although many of the comments were negative in sentiment, many contained constructive criticism (e.g., *"The video could be improved by providing more examples to better illustrate (and for the learner to better grasp) the difference between* *the various concepts: energy source, energy vector, secondary vectors, etc."*). A qualitative thematic analysis identified five common themes in the open-ended responses listed in Figure 3, which serves as a visual aid demonstrating that each of the identified themes were clearly present in both the experimental and control conditions.

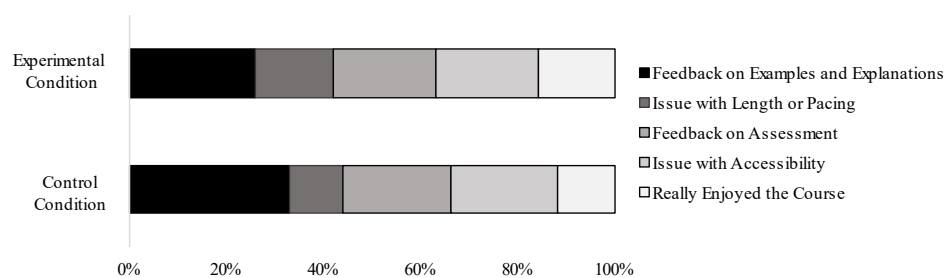

**Fig. 3.** Proportion of open-ended responses for each of the five major themes broken down by condition.

The most common theme, demonstrated in the quote given above, was learners wanting more examples and more detailed explanations for concepts (26% of responses; 7% of the sample). However, another subsample of the participants felt that the course was already too long with the pace being slightly quick (16% of responses; 4% of the full sample). Comments like this one, *"After the MCQs [multiple choice questions] at the end, it would have been good to have feedback that suggested what I should do to solidify my understanding where there were gaps"*, made up 21% of responses (5% of the full sample) giving suggestions related to the pre- and post-learning assessments. Another 16% of responses (4% of the full sample) made comments about accessibility issues, such as *"I'm not sure if there was a transcript available for the video, there might have been one that wasn't obvious to me"*, and the same proportion of participants explicitly stated that they enjoyed the course (16% of responses; 4% of the full sample). These qualitative findings suggest that the general perceptions of the instructional video content were similar for participants who viewed the AI-generated synthetic video and participants who viewed the traditional instructional video.

## 4       Discussion

The current study aimed to investigate the impact of AI-generated synthetic videos on learning performance compared to traditional instructor videos in an online learning platform. Our results indicated significant improvement in both traditional video and AI-generated video conditions between pre- and post-learning assessments with no statistical difference in terms of learning gains. This study contributes to the growing body of research on the use of synthetic characters and avatars in education by specifically focusing on the use of AI-generated synthetic videos and comparing them to traditionally produced instructor videos. However, it is important to note that AI-generated synthetic videos, which are created using generative AI methods and are not human-made, represent a new approach that differs from the use of human-made pedagogical agents and avatars. Interestingly, while our study shows an equal acceptance of the control and experimental conditions, research on pedagogical agents and avatars has generally shown that their use leads to more positive learning outcomes [15, 17, 19, 21]. It is possible that the neutral effect of these AI-generated synthetic videos compared to pedagogical agents and avatars is because the former is explicitly intended to replicate the traditional talking head video format. In contrast, the literature around pedagogical agents and avatars has focused on improving the learner experience by moving away from the talking head format to include meaningful gestures, and emotional cues [e.g., 18, 22].

In addition, the findings from both our qualitative and quantitative results examining the differences between AI-generated synthetic videos and traditionally produced instructor videos for learners in an online educational setting, suggest that there was little to no difference between learner perceptions of the two videos. Learner responses to our close-ended ended questions were nearly identical on three of the four items. Learner responses to our open-end question tended to be negative in sentiment, but this was not surprising given the wording of the prompt asking for suggestions to improve the course. We noted that sometimes the themes conflicted with one another. For example, the theme of learners wanting more examples and longer/more detailed explanations for concepts is at odds with the theme of the course being too long and the pace being slightly quick. Most of the themes were logistical and gave insight into how the micro-learning course could be improved from an accessibility standpoint. Organizing the findings from the thematic analyses by the type of video watched highlighted that there were no themes that were unique to either condition. The only observed difference in response to our close-ended items was in the extent to which the instructional video met the learners' expectations. The pattern of results showed that in the synthetic video with the AI-generated virtual instructor, more learners strongly agreed with the statement than stayed neutral. However, in the video with the traditional instructor, more learners stayed neutral than strongly agreed. While this difference in patterns between the conditions was not significant with the current sample size, the effect size indicator suggests a medium effect that warrants further investigations into potential explanations for any differences seen with larger sample sizes.

Having presented the value of AI-generated synthetic media, it is important to highlight that one of the greatest challenges of using generative AI in education is the fear of false or inaccurate information being presented to learners [13]. This is a valid concern as AI-generated content can be based on biased or incomplete data, which can lead to the dissemination of false or misleading information. However, in the current study, we addressed this challenge by using input data that was unchanged and came directly from material produced in collaboration with subject matter experts. That is, while the virtual instructor and video medium itself were synthetic in our experimental condition, the content of the media was not. Through the involvement of subject matter experts in the creation of the traditionally produced video, we ensured that the information presented in our AI-generated synthetic video was accurate and reliable. Furthermore, the utilization of AI-generated synthetic videos alone may not be sufficient to support effective learning. In order to fully leverage the potential of such videos in an educational setting, it is imperative to integrate them within a larger curriculum that is grounded in sound learning science and instructional design principles [16]. This holistic approach will ensure that the use of AI-generated synthetic videos is integrated in a manner that supports the overall learning objectives and outcomes, in conjunction with other pedagogical techniques such as formative and summative assessments, interactive activities, and opportunities for application and practice.

**Limitations.** Due to the desire for brevity in the micro-learning session context we studied, our pre- and post- learning assessments are brief. This limits our ability to investigate how AI-generated synthetic videos might differentially support various aspects of learning (e.g., difficulty of the material or complexity of concept) or what types of material they are most beneficial for (i.e., introductory material vs. advanced material). Additionally, our relatively small sample size may lead us to be underpowered to detect more nuanced differences in how learners perceive the two videos. This limits our ability to run more complex models to predict learning gains, including the investigation of fit between the learner and virtual instructor's demographics.

Future studies are warranted with a more robust learning assessment across several subject matters using larger sample sizes. This will allow for a deeper dive into how AI-generated virtual instructors are perceived differently than traditional instructors, and for what types of concepts they are most effective. Additionally, these types of designs would allow for exploration into whether human learning from AI-generated virtual instructors could be improved further by applying the research done with pedagogical agents. Future studies should consider factors such as learner preference at a larger scale, the use of AI-generated synthetic videos in longer duration learning paths, and the integration of more advanced generative AI techniques for improved quality to be evaluated in blind studies.

## 5   Conclusion

The adoption of generative AI to create synthetic instructional videos has the potential to be a viable substitute for videos produced via traditional methods in online educational settings. In terms of cost and time efficiency, the AI-generated synthetic video method is highly advantageous. The cost of production is near zero, while the traditional video required hours of human labor, film equipment, and software to produce. Additionally, the time to produce the synthetic video took only minutes or hours, as compared to the multiple hours or days required for the traditional video. Furthermore, updating or correcting errors in the original video would require a new round of filming and editing, while the AI-generated synthetic video method only requires editing the text script input and generating a new video, which can be done in minutes. These advantages can help us deliver high quality educational content for all learners across the globe. The current study is the first to indicate that learners have equal gains and learning experiences with an AI-generated virtual instructor as they do with a traditional instructor.

**References**

1. VanLehn, K., Banerjee, C., Milner, F., & Wetzel, J. (2020). Teaching Algebraic model construction: a tutoring system, lessons learned and an evaluation. *International Journal of Artificial Intelligence in Education*, *30*(3), 459-480. doi: 10.1007/s40593-020-00205-3
2. D'Mello, S., Picard, R. W., & Graesser, A. (2007). Toward an affect-sensitive AutoTutor. *IEEE Intelligent Systems, 22*(4), 53-61. doi: 10.1109/MIS.2007.79
3. Azevedo, R., Cromley, J. G., & Seibert, D. (2004). Does adaptive scaffolding facilitate students' ability to regulate their learning with hypermedia?. *Contemporary educational psychology*, *29*(3), 344-370. doi: 10.1016/j.cedpsych.2003.09.002
4. Luckin, R., Holmes, W., Griffiths, M., & Forcier, L. B. (2016). Intelligence unleashed: An argument for AI in education. Pearson Education, London.
5. Goodfellow, I., Pouget-Abadie, J., Mirza, M., Xu, B., Warde-Farley, D., Ozair, S., Courville, A., & Bengio, Y. (2014). Generative adversarial nets. *Advances in Neural Information Processing Systems 27,* 2672-2680. doi: 10.48550/arXiv.1406.2661.
6. Goodfellow, I., Pouget-Abadie, J., Mirza, M., Xu, B., Warde-Farley, D., Ozair, S., ... & Bengio, Y. (2020). Generative adversarial networks. *Communications of the ACM, 63*(11), 139-144. doi: 10.1145/3422622
7. Isola, P., Zhu, J.-Y., Zhou, T. & Efros, A. A. (2017). Image-to-image translation with conditional adversarial networks. *Proceedings of the IEEE Conference on Computer Vision and Pattern Recognition,* 1125–1134. doi: 10.1109/cvpr.2017.632
8. Zhang, Y., Weiss, R.J., Zen, H., Wu, Y., Chen, Z., Skerry-Ryan, R., Jia, Y., Rosenberg, A., Ramabhadran, B. (2019). Learning to speak fluently in a foreign language: Multilingual speech synthesis and cross-language voice cloning. *Proc. Interspeech*, 2080-2084. doi: 10.21437/Interspeech.2019-2668
9. Karras, T., Laine, S., Aittala, M., Hellsten, J., Lehtinen, J., & Aila, T. (2020). Analyzing and improving the image quality of StyleGAN. *Proc. IEEE/CVF Conference on Computer Vision and Pattern Recognition,* 8110-8119. doi: 10.1109/cvpr42600.2020.00813
10. Mirsky, Y. & Lee, W. (2012). The creation and detection of deepfakes: A survey. *ACM Computing Surveys, 54*, 1-41. doi: 10.1145/3425780
11. Radford, A., Wu, J., Child, R., Luan, D., Amodei, D., & Sutskever, I. (2019). Language models are unsupervised multitask learners. *OpenAI blog*, *1*(8), 9.
12. Brown, T., Mann, B., Ryder, N., Subbiah, M., Kaplan, J. D., Dhariwal, P., ... & Amodei, D. (2020). Language models are few-shot learners. *Advances in neural information processing systems*, *33*, 1877-1901.
13. Whittaker, L., Letheren, K., & Mulcahy, R. (2021). The rise of deepfakes: A conceptual framework and research agenda formMarketing. *Australasian Marketing Journal, 29*(3), 204-214. doi: 10.1177/1839334921999479
14. Mayer, R. E. (2003). The promise of multimedia learning: using the same instructional design methods across different media. *Learning and instruction*, *13*(2), 125-139. doi: 10.1016/s0959-4752(02)00016-6
15. Mayer, R. E. (2005). Cognitive theory of multimedia learning. *The Cambridge handbook of multimedia learning*, *41*, 31-48. doi: 10.1017/cbo9780511816819.004
16. Clark, R. C., & Mayer, R. E. (2016). E-learning and the science of instruction: Proven guidelines for consumers and designers of multimedia learning. John Wiley & Sons, New York. doi: 10.1002/9781119239086
17. Mabanza, N., & de Wet, L. (2014). Determining the usability effect of pedagogical interface agents on adult computer literacy training. In *E-Learning Paradigms and Applications* ( 145-183). Springer, Berlin. doi: 10.1007/978-3-642-41965-2_6

18. Lawson, A. P., Mayer, R. E., Adamo-Villani, N., Benes, B., Lei, X., & Cheng, J. (2021). Do learners recognize and relate to the emotions displayed by virtual instructors?. *International Journal of Artificial Intelligence in Education*, *31*(1), 134-153. doi:10.1007/s40593-021-00238-2
19. Schroeder, N. L., Adesope, O. O., & Gilbert, R. B. (2013). How effective are pedagogical agents for learning? A meta-analytic review. *Journal of Educational Computing Research, 49*(1), 1–39. doi: 10.2190/EC.49.1.a
20. Mayer, R. E. ,& DaPra, C. S. (2012). An embodiment effect in computer-based learning with animated pedagogical agents. *Journal of Experimental Psychology: Applied, 18*, 239–252. doi: 10.1037/a0028616
21. Wang, F., Li, W., Xie, H., & Liu, H. (2017). Is a pedagogical agent in multimedia learning good for learning? A meta-analysis. *Advances in Psychological Science*, *25*(1), 12. doi: 10.3724/SP.J.1042.2017.00012
22. Wang, F. X., Li, W. J., Mayer, R. E., & Liu, H. S. (2018). Animated pedagogical agents as aids in multimedia learning: Effects on eye fixations during learning and learning outcomes. *Journal of Educational Psychology, 110,* 250–268. doi: 10.1037/edu0000221
23. Hudson, I. & Hurter, J. Avatar types matter: Review of avatar literature for performance purposes. In *Proc. International Conference on Virtual, Augmented and Mixed Reality (*14-21). Springer, Cham. doi: 10.1007/978-3-319-39907-2_2
24. Pataranutaporn, P., Danry, V., Leong, J., Punpongsanon, P., Novy, D., Maes, P., & Sra, M. (2021). AI-generated characters for supporting personalized learning and well-being. *Nature Machine Intelligence*, *3*(12), 1013-1022. doi: 10.1038/s42256-021-00417-9
25. Pataranutaporn, P., Leong, J., Danry, V., Lawson, A. P., Maes, P., & Sra, M. (2022). AI-generated virtual instructors based on liked or admired people can improve motivation and foster positive emotions for learning. *IEEE Frontiers in Education Conference* (1-9). doi: 10.1109/FIE56618.2022.9962478.